\documentclass[letterpaper, 10 pt, conference]{ieeeconf}  

\usepackage{xspace}
\usepackage{amsfonts}		
\usepackage[subrefformat=parens]{subcaption}		
\usepackage{siunitx}		
\usepackage{acronym}		

\usepackage{amsmath} 
\usepackage{multirow}
\usepackage{nicefrac}       
\usepackage{algorithm}
\usepackage{algpseudocode}
\usepackage{makecell}		
\usepackage{pifont}			

\usepackage{booktabs} 

\usepackage{graphics}
\graphicspath{{./figures/}}

\usepackage{tikz}
\usepackage{pgfplots}
\usepackage{pgfplotstable}
\usepackage{tikzscale}

\pgfplotsset{compat=1.16}

\pgfplotstableset{search path={.,./figures,./measurements,../measurements}}

\usetikzlibrary{external}
\usetikzlibrary{positioning}			
\usetikzlibrary{calc}					
\usetikzlibrary{arrows}
\usetikzlibrary{decorations.markings}
\usetikzlibrary{shapes}
\usetikzlibrary{spy}
\usetikzlibrary{pgfplots.groupplots}
\usetikzlibrary{patterns}
\usetikzlibrary{patterns.meta}

\usepgfplotslibrary{fillbetween}

\usepgfplotslibrary{colorbrewer}
\pgfplotsset{
    cycle list/Set1-5,
    cycle multiindex* list={
        mark list*\nextlist
        Set1-5\nextlist,
    },
}

\usepackage{ifthen}

\newboolean{symcheck}
\setboolean{symcheck}{false}

\newboolean{shownotes}
\setboolean{shownotes}{false}

\newboolean{dosavespace}
\setboolean{dosavespace}{false}

\ifthenelse{\boolean{symcheck}}{
	\NewDocumentCommand{\symcheck}{m e{_^}}{%
		{\color{blue!60!black}%
			\IfNoValueTF{#2}{%
				\IfNoValueTF{#3}{#1}{#1^{\color{black}#3}}%
			}{%
				\IfNoValueTF{#3}{#1_{\color{black}#2}}{#1_{{\color{black}#2}}^{\color{black}#3}}%
		}}%
	}
}{
	\newcommand*{\symcheck}[1]{#1}
}

\ifthenelse{\boolean{dosavespace}}{
	
}{

}

\ifthenelse{\boolean{shownotes}}
{
	\usepackage[textsize=footnotesize,linecolor=red,bordercolor=red,backgroundcolor=red!10]{todonotes}	
	\usepackage{setspace}
	\usetikzlibrary{external}
	\setlength{\marginparsep}{2pt}
	\setlength{\marginparwidth}{17.5mm}

	\newcommand{\note}[4]{%
		\ifthenelse{\equal{#1}{footnote}}{{
			\tikzexternaldisable%
			\footnote{#2\todo[caption={},#3,inline]{#4}}%
		}}{{
			\tikzexternaldisable%
			\todo[caption={},#3,#1]{#4}%
		}}{}%
	}

	\newcommand{\draft}[1]{{
		\color{black!30} \vspace{1ex}\hrule\vspace{2pt}
		\color{blue!55!black} #1
		\color{black!30} \vspace{2pt}\hrule\vspace{1ex}
	}}

}
{
	\usepackage[disable]{todonotes}

	\newcommand{\note}[4]{}
	\newcommand{\draft}[1]{}
}

\usepackage{cite}
\usepackage[hidelinks]{hyperref}
\hypersetup{
		colorlinks=true,
		urlcolor=blue,
		linkcolor=blue,
		citecolor=black,
	}
\usepackage[capitalise]{cleveref}		

\newcommand{\todox}   [2][]{\note{#1}{\color{red}}{linecolor=red,bordercolor=red,backgroundcolor=red!10}{TODO: #2}}

\newcommand{\todoa}[3]{\todox[#1]{#3 [#2]}}

\newcommand{\todoag}[2][]{\todoa{#1}{A.}{#2}}

\newcommand{\fakepar}[1]{\textbf{#1.}}  

\newcommand{\capt}[1]{\mdseries{\textit{#1}}}

\crefname{section}{Sec.}{Secs.}
\Crefname{section}{Section}{Sections}
\crefname{figure}{Fig.}{Figures}
\Crefname{figure}{Figure}{Figures}

\newcommand*{\eg}{e.g.,\xspace}

\newcommand*{\etc}{etc.\xspace}

\DeclareSIUnit{\belmilliwatt}{Bm}
\DeclareSIUnit{\dBm}{\deci\belmilliwatt}

\acrodef{name}[\textsc{Flare}]{\symcheck{Federated Learning Architecture for Resilient Embedded Learning}}
\acused{name}

\acrodef{cf}[CF]{Crazyflie}
\acrodefplural{cf}[CF quadcopters]{Crazyflie quadcopters}

\acrodef{cf2}[CF 2.x]{Crazyflie 2.x}
\acrodefplural{cf}[CF 2.x quadcopters]{Crazyflie quadcopters}
\acused{cf2}

\acrodef{cfb}[CFB]{Crazyflie 2.1 Brushless}		
\acrodefplural{cfb}[CB quadcopters]{Crazyflie 2.1 Brushless quadcopters}  

\acrodef{nanoquad}[nano-quadcopter]{nano-quadcopter}
\acrodefplural{nanoquad}[nano-quadcopters]{nano-quadcopters}
\acused{nanoquad}

\acrodef{esc}[ESC]{electric speed controller}
\acrodefplural{esc}[ESCs]{electric speed controllers}

\acrodef{mjx}[MJX]{MuJoCo XLA}

\acrodef{jax}[JAX]{JAX}
\acused{jax}

\acrodef{imu}[IMU]{inertial measurement unit}
\acrodef{mocap}[motion capture system]{motion capture system}
\acused{mocap}

\acrodef{ble}[BLE]{Bluetooth Low Energy}
\acrodef{rl}[RL]{reinforcement learning}
\acrodef{ppo}[PPO]{proximal policy optimization}
\acrodef{bo}[BO]{bayesian optimization}
\acrodef{nn}[NN]{neural network}
\acrodef{mpc}[MPC]{model predictive control}

\acrodef{srt}[SRT]{single-rotor thrust}

\acrodef{rpm}[RPM]{revolutions per minute}

\newcommand{\opindent}{\phantom{=}}
\newcommand{\realnmbrs}{\ensuremath{\mathbb{R}}}

\newcommand{\forcemotor}{\ensuremath{F_\mathrm{m}}}
\newcommand{\forcethrust}{\ensuremath{F_\mathrm{thrust}}}
\newcommand{\forceaero}{\ensuremath{F_\mathrm{a}}}
\newcommand{\taumotor}{\ensuremath{\tau_\mathrm{m}}}
\newcommand{\tauthrust}{\ensuremath{\tau_\mathrm{f}}}

\newcommand{\jmotor}{\ensuremath{J_\mathrm{m}}}  
\newcommand{\twonorm}[1]{||#1||_\mathrm{2}}

\newcommand{\flipcontroller}{flip-controller}

\IEEEoverridecommandlockouts                              

\usepackage{fancyhdr}

\title{\LARGE \bf
How to Model Your Crazyflie Brushless
}

\newif\ifanonymize

\ifanonymize
        \author{Anonymous Authors}
\else
        \author{Alexander Gräfe$^{1}$, Christoph Scherer$^{2}$, Wolfgang Hönig$^{2}$ and Sebastian Trimpe$^{1}$
        \thanks{This work has been supported by the German Federal Ministry of Research, Technology and Space (BMFTR) under the Robotics Institute Germany (RIG). The authors gratefully acknowledge the computing time provided to them at the NHR Center NHR4CES at RWTH Aachen University (project number p0021919). This is funded by the Federal Ministry of Education and Research, and the state governments participating on the basis of the resolutions of the GWK for national high performance computing at universities (www.nhr-verein.de/unsere-partner).}%
        \thanks{$^{1}$Alexander Gräfe and Sebastian Trimpe are with the Institute for Data Science in Mechanical Engineering,
                                        RWTH Aachen University, Germany
                                        {\tt\small alexander.graefe@dsme.rwth-aachen.de, trimpe@dsme.rwth-aachen.de}}%
        \thanks{$^{2}$Christoph Scherer and Wolfgang Hönig are with the Intelligent Multi-Robot Coordination Lab, TU Berlin, Germany
                                        {\tt\small c.scherer@tu-berlin.de, hoenig@tu-berlin.de}}%
        }
\fi

\makeatletter
\let\NAT@parse\undefined
\makeatother

\begin{document}

{\onecolumn \begin{center} Accepted for publication at the 2026 IEEE International Conference on Robotics \& Automation.\end{center}
\noindent\fbox{%
        \parbox{\textwidth}{%
                © 2026 IEEE. Personal use of this material is permitted. Permission from IEEE must be obtained for all other uses, in any current or future media, including reprinting/republishing this material for advertising or promotional purposes, creating new collective works, for resale or redistribution to servers or lists, or reuse of any copyrighted component of this work in other works.
        }%
}
}
\twocolumn
\newpage

\maketitle
\pagestyle{empty}

\begin{abstract}
The Crazyflie quadcopter is widely recognized as a leading platform for nano-quadcopter research. 
In early 2025, the Crazyflie Brushless was introduced, featuring brushless motors that provide around 50\% more thrust compared to the brushed motors of its predecessor, the Crazyflie 2.1.
This advancement has opened new opportunities for research in agile nano-quadcopter control.
To support researchers utilizing this new platform, this work presents a dynamics model of the Crazyflie Brushless and identifies its key parameters. 
Through simulations and hardware analyses, we assess the accuracy of our model.
We furthermore demonstrate its suitability for reinforcement learning applications by training an end-to-end neural network position controller and learning a backflip controller capable of executing two complete rotations with a vertical movement of just 1.8 meters.
This showcases the model's ability to facilitate the learning of controllers and acrobatic maneuvers that successfully transfer from simulation to hardware.
Utilizing this application, we investigate the impact of domain randomization on control performance, offering valuable insights into bridging the sim-to-real gap with the presented model.
We have open-sourced the entire project, enabling users of the Crazyflie Brushless to swiftly implement and test their own controllers on an accurate simulation platform.

\end{abstract}

\section{INTRODUCTION}
Over the past decade, \ac{cf} quadcopters have become state-of-the-art in indoor \ac{nanoquad} research and teaching, both for single quadcopters and quadcopter swarms~\cite{preiss2017crazyswarm,giernacki_crazyflie_2017,budaciu_evaluation_2019,socas_control_2021,wahba_kinodynamic_2024,grafe2025dmpc} \todoag{Add refs}. 
At the beginning of 2025, a new version, the \ac{cfb}, was released.
In contrast to the original \ac{cf} 2.1 quadcopter that features brushed motors, the \ac{cfb} is equipped with brushless motors and \acp{esc}.
Consequently, the \ac{cfb} boasts a significantly higher thrust-to-weight ratio of around 3:1 compared to the 2:1 ratio of the \ac{cf} 2.1. This increase in power positions the \ac{cfb} as a valuable platform for research and teaching focused on agile control of \acp{nanoquad}.

To facilitate method development involving the \ac{cfb}, this work derives a dynamics model of the \ac{cfb}. 
We identify key parameters such as thrust curves and motor time constants and provide guidelines for manually re-identifying parameters like moments of inertia, which become necessary when modifications are made to the quadcopter's setup.
Furthermore, to demonstrate the model's effectiveness for control design, we employ it to train two end-to-end \ac{nn} controllers entirely in simulation based on the identified model using \ac{rl}. 
The first controller is capable of performing position control, which is comparable to the controllers embedded in the \ac{cfb}'s firmware.
The second controller can execute up to two consecutive backflips with a vertical movement of \SI{1.8}{\meter}.
We use this application of our model to assess the extent of necessary domain randomization, offering valuable insights into the sim-to-real gap.

In summary, our contributions are:
\begin{enumerate}
    \item We derive a dynamics model of the \ac{cfb}, including motor dynamics, and offer practical guidelines for manual parameter identification.
    \item We evaluate the model in two ways: (a) by assessing its prediction accuracy and (b) by training end-to-end \ac{nn} controllers for the \ac{cfb}, deploying it on real Crazyflie Brushless hardware, and using it to analyze critical choices such as domain randomization. 
\end{enumerate}

\sloppy{The codebase, including the model, its parameters and a simulator based on \ac{jax}~\cite{jax2018github} and \ac{mjx}~\cite{todorov2012mujoco} for highly parallelized simulations on GPUs, is available under \href{https://github.com/Data-Science-in-Mechanical-Engineering/CrazyflieBrushJAX}{github.com/Data-Science-in-Mechanical-Engineering/CrazyflieBrushJAX}.
Additionally, an accompanying video demonstrating the hardware experiments is available at \href{http://tiny.cc/CFBVideo}{tiny.cc/CFBVideo}.}

The remainder of this work is organized as follows. 
First, we review related research on modeling previous versions of the \ac{cf} in \cref{sec:related_work} and give an overview of the components of the \ac{cf} Brushless in \cref{sec:background}.
Next, we derive the model and identify its parameters in \cref{sec:model}. 
Subsequently, we evaluate the model's predictive performance in \cref{sec:evaluation_model} and conduct experiments using end-to-end \ac{rl} to explore the sim-to-real gap in \cref{sec:evaluation_rl}.
\section{RELATED WORK}
\label{sec:related_work}
This section reviews related work on modeling previous versions of the \ac{cf}, referred to as \ac{cf2}. 
Additionally, it presents application for \ac{cf} models, emphasizing the relevance of the \ac{cf} platform in robotics research.

\subsection{Crazyflie Modeling}
\label{sec:related_work:model}
Quadcopter dynamics are generally well understood and models feature varying levels of detail. 
These range from simple models that treat the quadcopter as a rigid body subject to external forces and torques to more complex models incorporating learned residual dynamics and even full fidelity fluid simulators~\cite{Mueller2025, bauersfeld2021neurobem, kaufmann2023champion, paz2021assessment, yoon2017computational}.

For the \ac{cf2}, several studies have derived models all with similar levels of detail~\cite{forster2015system, landry2015planning, panerati2021learning,eschmann2024learning}.
In these models, the \ac{cf} is represented as a rigid body influenced by motor forces and reactive torques. 
The motor forces and torques are either modeled using static functions~\cite{panerati2021learning, landry2015planning} or combined with a dynamic system~\cite{forster2015system, eschmann2024learning}, where motor commands are processed through a first-order dynamic system to simulate motor dynamics.

Building on these models, several simulators have been developed. 
We highlight three of the most common and recent ones: 
PyBullet-drones~\cite{panerati2021learning}, which utilizes PyBullet~\cite{coumans2021} to simulate multiple quadcopters for swarming simulations with vision support; 
CrazySim~\cite{LlanesICRA2024}, based on Gazebo~\cite{Koenig2004}, offering software-in-the-loop simulations via \ac{cf} firmware wrappers; 
and Crazyflow~\cite{crazyflow}, currently under development for highly parallelized simulations of \ac{cf} swarms using \ac{mjx}.

The presented models and simulators focus on the \ac{cf2} and to our knowledge, no comparable dynamics model of the new \ac{cf} is openly accessible.
To bridge this gap, this work introduces both a model and simulator specifically for the \ac{cfb}. 
We maintain a similar level of detail as existing models, while basing our simulator on \ac{mjx}, akin to Crazyflow~\cite{crazyflow}.

\subsection{Crazyflie Model Use Cases}
Existing \ac{cf} models have found extensive application in domains such as control design and multi-robot systems. 
For instance, Socas et al.~\cite{socas_control_2021} used the model to design and evaluate periodic and event-based PID controllers, while Nguyen et al.~\cite{nguyen2024tinympc} leveraged a linearized model to develop an onboard \ac{mpc} for dynamic flight. 
For multi-robot systems, Gräfe et al.~\cite{Graefe2022,grafe2025dmpc} evaluated a distributed \ac{mpc} for swarms using PyBullet-drones~\cite{panerati2021learning} and Wahba et al.~\cite{wahba_efficient_2024,wahba_kinodynamic_2024} used the model to design collaborative transport algorithms of cable-suspended payloads.

The models have also been widely adopted for \ac{rl}. 
Applications use \ac{rl} to optimize PID gains~\cite{javeed2023reinforcement}, manage swarms with high-level commands via multi-agent \ac{rl}~\cite{felten2024} and learn end-to-end control policies~\cite{do2024end,eschmann2024learning}.

To demonstrate the capability of the \ac{cfb} model proposed in this work, we use it to train an end-to-end \ac{nn} controller with \ac{rl}. 
This application serves as a test of model precision, as end-to-end learning is highly sensitive to model inaccuracies~\cite{kaufmann2022benchmark, ferede2024end}. 
Furthermore, it enables an evaluation of the domain randomization required for successful sim-to-real transfer, which serves as a metric for the model's accuracy.

\section{BACKGROUND: CRAZYFLIE BRUSHLESS}
\label{sec:background}

\begin{figure}
    \centering
    \vspace{3mm}
    \includegraphics[width=0.48\textwidth]{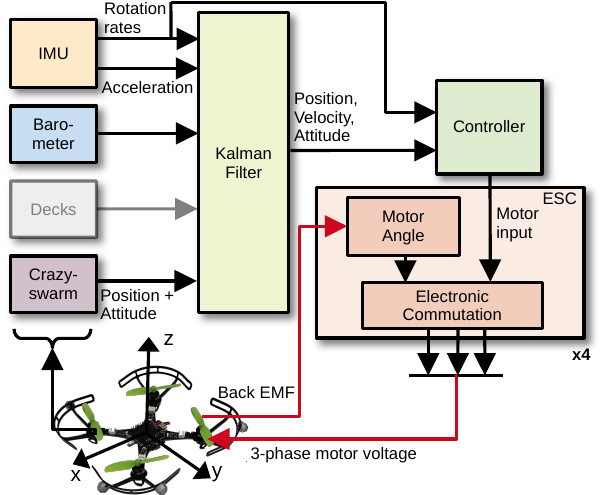}
    \caption{Overview of the different components of the \ac{cfb}~\cite{bitcraze2025crazyflie,crazyflie_bl_schematics}. \capt{A Kalman filter estimates position, velocities and attitude from the data of different sensors. The controller uses these in combination with rotation rates of the \ac{imu} to give motor inputs to the \acp{esc}, which perform the electronic commutation of the brushless motors using the motor angle measured via the back electromotive force (EMF).}}
    \label{fig:cfb_overview}
\end{figure}

This section provides an overview of the various components of the \ac{cfb} as depicted in \cref{fig:cfb_overview}.

\fakepar{Hardware} In the largest parts, the \ac{cf2} and \ac{cfb} are identical~\cite{crazyflie_schematics,crazyflie_bl_schematics}.
They comprise two processors, one dedicated to \SI{2.4}{\giga\hertz} communication and an application processor (STM32F405 ARM Cortex-M4 single core MCU with 168MHz, \qty{192}{\kilo\byte} SRAM and \qty{1}{\mega\byte} flash) for processing tasks.
Additionally, they feature an \ac{imu} equipped with a 3-axis accelerometer and gyroscope (BMI088), along with a pressure sensor (BMP388). 
They can be expanded using various decks that provide additional sensors.

The primary difference between the \ac{cf2} and the \ac{cfb} lies in their motor systems.
The \ac{cf2} utilizes brushed motors, which are controlled by a transistor.
In contrast, the \ac{cfb} employs four \acp{esc} to control its brushless motors.
Each \ac{esc} consists of an EFM8BB21 MCU and three half-bridges.

\fakepar{Software}
The application processor manages the entire state estimation, control and planning pipeline. 
It employs an extended Kalman filter~\cite{julier1997new, mueller2015fusing,mueller2017covariance}, operating at \SI{100}{\hertz}, to fuse the measurements from the \ac{imu}, pressure sensor, other decks and, if available, \ac{mocap} information into a state estimate.
Subsequently, the controller computes motor commands which are proportional to the motor voltage.
These commands are transmitted to four \acp{esc}, which perform the electronic commutation and apply the desired motor voltage.
For this, each \ac{esc} measures the motor's angle using the induced voltage of the rotating motor.
It then computes and applies the three-phase motor voltages correspondingly.

\section{SYSTEM MODEL}
\label{sec:model}

In this section, we first present the dynamics model in \cref{sec:model:formula} and then its parameters in \cref{sec:model:sysid}. 
Finally, in \cref{sec:model:simulator}, we give an overview of our simulator implementation based on \ac{mjx}.

\subsection{First-principle model}
\label{sec:model:formula}
 
When modeling the \ac{cfb}, we preserve the same level of detail as existing models for the \ac{cf} 2.x~\cite{forster2015system,landry2015planning,LlanesICRA2024,panerati2021learning,eschmann2024learning} (cf. \cref{sec:related_work:model}). 
In \cref{sec:evaluation_model}, we enumerate unmodeled effects that could become significant in research exploring the limits of \ac{cfb} dynamics.

First, we present the comprehensive set of equations that describe the quadcopter's dynamics. 
Subsequently, we will examine each equation and its parameters in a step-by-step manner. 
Our model follows common quadcopter models presented, e.g., in works~\cite{bauersfeld2021neurobem,kaufmann2023champion, ferede2024end,ferede2024end2,Mueller2025,forster2015system,landry2015planning,LlanesICRA2024,panerati2021learning,eschmann2024learning}.
We describe the \ac{cfb} using a state-space model $\dot{x}=f(x, u)$, where $x\in\mathbb{R}^{17}$ is the state and $u\in[0,1]^{4\times1}$ are the motor commands.
In detail, the model is
\newcounter{mymatrow}
\stepcounter{mymatrow}
\begin{equation*}
\dot{x} = \begin{bmatrix}
\dot{p} \\
\dot{v} \\
\dot{q} \\
\dot{\omega} \\
\dot{\Omega}
\end{bmatrix}
=
\begin{bmatrix}
v \\
\frac{1}{m}\forcemotor(\Omega, q) - ge_\mathrm{z} \\  
\frac{1}{2} \, q \otimes \begin{bmatrix}0 & \omega\end{bmatrix}^T \\
J^{-1}(\taumotor(\Omega, u) - \omega \times (J\omega)) \\  
-\frac{1}{T}\Omega + \frac{K}{T}u
\end{bmatrix}\hspace{5pt}
\begin{array}{l}
\stepcounter{equation}(\theequation)\\
\stepcounter{equation}(\theequation)\\
\stepcounter{equation}(\theequation)\\
\stepcounter{equation}(\theequation)\\
\stepcounter{equation}(\theequation)\\
\end{array},
\end{equation*}
where $v\in\mathbb{R}^3$ is the velocity, $q\in S^3$ are the quaternions representing orientation, $\omega\in\mathbb{R}^3$ is the angular velocity in the body frame, $\Omega\in\mathbb{R}^4$ are the rotation speeds of the motors and $g=\SI{9,81}{\meter\per\square\second}$.
The coordinate system is depicted in \cref{fig:cfb_overview} and roll, pitch and yaw rates rotate around x, y and z axes respectively. 
These rotations are deemed positive when counter-clockwise. 
All parameters can be found in Table~\ref{tab:parameters}.
We now go through the equations explaining their parts.

\fakepar{Rotational and Translational Dynamics (1)--(4)}
The translational dynamics of the quadcopter are a standard model, summing up the influence of the motor's thrust (\forcemotor, \taumotor).
The thrust that the motors generate is a nonlinear function of the motor rotational speed
\begin{equation}
    \forcemotor(\Omega, q) = R(q)e_\mathrm{z}\begin{bmatrix}1, 1, 1, 1\end{bmatrix}\forcethrust(\Omega),
\end{equation}    
where $R(q)\in\realnmbrs^{3\times3}$ is the rotation matrix and $\forcethrust(\Omega):\realnmbrs^{4\times1}\to\realnmbrs^{4\times1}$ is an element-wise 3rd order polynomial with units $\SI{}{\radian\per\second}\to\SI{}{\newton}$ (cf. \cref{sec:model:sysid}).

The torques from the motors in pitch and roll direction are the thrusts of the motors times the orthogonal distance to the center of mass. 
The torque around the z-axis is the reactive torque of the motor, which is the sum of the inertia torque $\jmotor\dot{\Omega}
$ and the aerodynamic/friction torques on the motor $\tauthrust$
\begin{align}
    \taumotor(\Omega, u) &= \begin{bmatrix}L&0&0&0&0\\0&-1&1&-1&1\end{bmatrix}\begin{bmatrix}\forcethrust(\Omega) \\ \jmotor \dot{\Omega} + \tauthrust(\Omega)\end{bmatrix},
\end{align}
where we measured the element-wise 3rd order polynomial $\tauthrust(\Omega):\realnmbrs^4\to\realnmbrs^4$ with dimension $\SI{}{\radian\per\second}\to\SI{}{\newton\meter}$ (cf. \cref{sec:model:sysid})
and 
\begin{equation}
    L = \begin{bmatrix}-\ell/2&-\ell/2&\ell/2&\ell/2\\-\ell/2&\ell/2&\ell/2&-\ell/2\end{bmatrix},
\end{equation}
where $\ell$ is the width of the quadcopter.

While accounting for aerodynamic forces in the model could further enhance accuracy, accurately modeling these effects is challenging~\cite{bauersfeld2021neurobem,kaufmann2023champion,hanover2024autonomous}. 
Our current results indicate that effective performance in low speed scenarios ($\leq \SI{3}{\meter\per\second}$) can be achieved by solely modeling the rigid-body dynamics, motor dynamics and motor response.

\begin{table}[t]
    \vspace{3mm}
    \caption{Symbols, explanations and dimensions for the quadcopter model.}
    \label{tab:parameters}
    \fontsize{8}{10}\selectfont
    \centering
    \begin{tabular}{lll}
        \toprule
        Symbol & Explanation & Dimension \\
        \midrule
        $v$ & Linear velocity & $\realnmbrs^{3\times1}$ (\si{\meter\per\second}) \\
        $q$ & Quaternion & $S^3$ (unitless) \\  
        $\omega$ & Angular velocity in body frame & $\realnmbrs^{3\times1}$ (\si{\radian\per\second}) \\
        $\Omega$ & motor rotation speed & $\realnmbrs^{4\times1}$ (\si{\radian\per\second}) \\
        $u$ & Motor command & $[0,1]^{4\times1}$ (unitless) \\
        $m$ & Quadcopter mass & $\realnmbrs$ (\si{\kilo\gram}) \\
        $\ell$ & Quadcopter width & $\realnmbrs$ (\si{\meter}) \\
        $J$ & Quadcopter moment of inertia & $\realnmbrs^{3\times3}$ (\si{\kilo\gram\meter^2}) \\
        $T$ & motor time constant & $\realnmbrs$ (\si{\second}) \\
        $K$ & motor amplification factor & $\realnmbrs$ (\si{\radian\per\second}) \\
        $\jmotor$ & motor moment of inertia & $\realnmbrs$ (\si{\kilo\gram\meter^2}) \\
        \bottomrule
    \end{tabular}
\end{table}

\fakepar{Motor Dynamics (5)}
The motor dynamics model summarizes the behavior of the dynamics of the motor rotational mass, the coils and the \acp{esc}.
However, developing a model of all this proves challenging, especially given the complicated behavior of the \ac{esc}'s electronic commutation, rate limitation \etc{}
We hence approximate the dynamics via a first-order linear system with time constant $T$ and amplification $K$ as done in~\cite{bauersfeld2021neurobem,kaufmann2023champion, ferede2024end,ferede2024end2,eschmann2024learning}.

\input{figures/motor.tex}
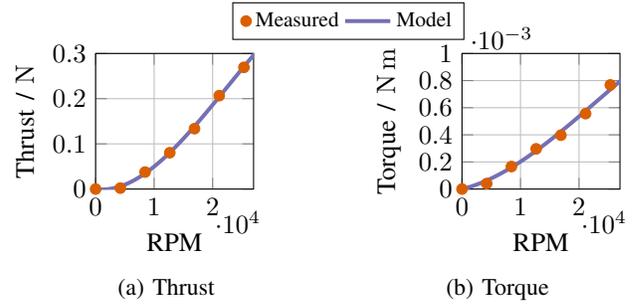
\begin{figure}[t]
    \newcommand{\plotwidth}{0.87\linewidth}
    \newcommand{\plotheight}{0.8\linewidth}
    \centering
    
    \begin{subfigure}[b]{0.49\linewidth}
        \centering
        \begin{tikzpicture}
            \begin{axis}[
                width=\plotwidth,
                height=\plotheight,
                xlabel={RPM},
                ylabel={Thrust / \si{\newton}},
                legend style={at={(1.6,1.1)}, anchor=south, legend columns=2,nodes={scale=0.8, transform shape}},
                grid=major,
                xmin=0, xmax=27000,
                ymin=0, ymax=0.3,
                x tick scale label style={yshift=10pt},
            ]
            
            \addplot[
                color=Dark2-B,
                mark=*,
                mark size=2pt,
                only marks,
            ] table[
                x=omega_thrust,
                y=measured_thrust,
                col sep=comma
            ] {../figure_data/thrust_curve_measured.csv};
            \addlegendentry{Measured}
            
            \addplot[
                color=Dark2-C,
                mark=none,
                line width=1.5pt,
            ] table[
                x=omega_theo,
                y=theo_thrust,
                col sep=comma
            ] {../figure_data/thrust_curve_model.csv};
            \addlegendentry{Model}
            
            \end{axis}
        \end{tikzpicture}
        \caption{Thrust}
        \label{fig:thrust}
    \end{subfigure}
    \hfill
    \begin{subfigure}[b]{0.49\linewidth}
        \centering
        \begin{tikzpicture}
            \begin{axis}[
                width=\plotwidth,
                height=\plotheight,
                xlabel={RPM},
                ylabel={Torque / \si{\newton\meter}},
                grid=major,
                xmin=0, xmax=27000,
                ymin=0, ymax=0.001,
                x tick scale label style={yshift=10pt},
            ]
            
            \addplot[
                color=Dark2-B,
                mark=*,
                mark size=2pt,
                only marks,
            ] table[
                x=omega_torque,
                y=measured_torque,
                col sep=comma
            ] {../figure_data/thrust_curve_measured.csv};
            
            \addplot[
                color=Dark2-C,
                mark=none,
                line width=1.5pt,
            ] table[
                x=omega_theo,
                y=theo_torque,
                col sep=comma
            ] {../figure_data/thrust_curve_model.csv};
            
            \end{axis}
        \end{tikzpicture}
        \caption{Torque}
        \label{fig:torque}
    \end{subfigure}
    
    \caption{Comparison between measured and model-predicted thrust and torque characteristics. \capt{The propeller model accurately captures both thrust and torque behavior across the operating range.}}
    \label{fig:thrust_torque}
\end{figure}

\subsection{System identification}
\label{sec:model:sysid}
We identify the parameters of the model in three different experiments.
First, we measure the motor parameters with sensors measuring the motor's \ac{rpm}.
Then, we measure the thrust and reactive torques $\forceaero(\Omega)$ and $\forcethrust(\Omega)$ 
and finally identify the remaining parameters (moment of inertia matrix) by fitting measured trajectories of the quadcopter to simulated ones.

\fakepar{Motor Dynamics Identification}
We identify the motor system dynamics by recording the motor commands and the \ac{rpm} during flight.
While we can directly assess the motor commands in the firmware of the quadcopter, we measure the \ac{rpm} using a custom deck with a QRD1114 infrared sensor and a reflective band beneath the propeller as described in~\cite{crazyflie_pwm_to_thrust}.
We then fit the time constant $T$ and amplification $K$ to the measured curves leading to $T=\SI{0.05}{\second}$ and $K=\SI{2900}{\radian\per\second}$.

\fakepar{Thrust Measurements}
We place the quadcopter on a force/torque measurement testbed we built using a scale and a rotating arm.
We increase the motor command of only one motor in steps and once it has reached a steady state, we measure the force/torque.
Afterwards, we fit a polynomial through the points measured leading to
\begin{equation}
       \forcethrust(\Omega)= -0.23(\frac{\Omega}{\sigma})^3 + 0.562(\frac{\Omega}{\sigma})^2 -0.043(\frac{\Omega}{\sigma}),
\end{equation}
\begin{align}
\tauthrust(\Omega)&=10^{-4}\left[-3.4(\frac{\Omega}{\sigma})^3 +  8.7(\frac{\Omega}{\sigma})^2  +  2.9(\frac{\Omega}{\sigma})\right]
\end{align}
with $\sigma=2.9\cdot10^3$ for numerical stability.

\fakepar{Fitting Measured Trajectories}
All remaining parameters are measured by manually fitting simulated trajectories to measured ones (\cref{tab:parameter_value}).
We emphasize that especially the moments of inertia are dependent on the additional payload of the quadcopter, e.g., whether equipped with propeller guards, \ac{mocap} markers and decks.
The given ones should, however, give a good starting ground to quickly identify the parameters for the \ac{cfb} in a specific application using the following procedures on a flying quadcopter.

For the inertias around x and y axis, the quadcopter should perform a maneuver where it rotates around its x and y axes.
This can be achieved by swapping from a stabilizing controller to a policy that gives each motor different motor commands such that they create torques around the x and y axis. 
After a short amount of time (around \SI{500}{\milli\second}), the quadcopter swaps back to the stabilizing controller to avoid crashing.
This allows us to fit the moments of inertia around x and y axis by aligning the simulated and measured rotation rates around x and y axis, respectively.

Similarly, for the moment of inertia around z axis and $\jmotor$, we perform an experiment, where we keep the motor command of three motors low and give a high command to one, which causes the yaw rotation rate to rise fast.
This yaw rotation rate can be used to fit the parameters.

\begin{table}[t]
    \vspace{3mm}
    \caption{Symbols, explanations and dimensions for the quadcopter model. \capt{Values in parentheses are for the \ac{cfb} without propeller guards.}}
    \label{tab:parameter_value}
    \fontsize{8}{10}\selectfont
    \centering
    \begin{tabular}{ll}
        \toprule
        Symbol & Value \\
        \midrule
        $m$ & \SI{44}{\gram} (\SI{40}{\gram}) \\
        $J$ & \makecell[l]{$\text{diag}([3.3, 3.6, 5.9])\cdot10^{-5} \si{\kilo\gram\meter^2}$\\($\text{diag}([1.8, 2.4, 3])\cdot10^{-5}$ \si{\kilo\gram\meter^2})} \\
        $T$ & $\SI{50}{\milli\second}$ \\
        $K$ & $2.9\cdot10^3\si{\radian\per\second}$ \\
        $\jmotor$ & $0.5\cdot10^{-7}\si{\kilo\gram\meter^2}$\\
        $\ell$ & \SI{7.07}{\centi\meter}\\
        \bottomrule
    \end{tabular}
\end{table}

\subsection{\ac{mjx} Simulator}
\label{sec:model:simulator}

We implement the derived model inside \ac{mjx}~\cite{todorov2012mujoco}. 
Although the differential equation of the drone could also be solved using a suitable solver, \ac{mjx} allows to reuse the simulator for complex tasks that involve \eg contacts.
MuJoCo itself takes care of the translational and rotational dynamics.
We place the motor model outside of MuJoCo and solve it separately at each step.
As the dynamics are linear, this can be done exactly.
From $\Omega$, we then calculate the thrust reactive torque of the motor and give this to the MuJoCo model as input.
We simplify the dynamics of the Kalman filter, the sensor low-pass filters and the delay of communication with the \ac{mocap} via Crazyswarm with a fixed delay of \SI{8}{\milli\second}.

\section{EVALUATION: MODEL ACCURACY}
\label{sec:evaluation_model}
This section evaluates the predictive capabilities of the model by comparing simulated and measured trajectories.

\subsection{Method}
We will evaluate the model on two different maneuvers flown by the \ac{cfb}.
First, a maneuver, where the \ac{cfb} changes from hovering to a horizontal flight.
And second, a maneuver, where we disturb the yaw orientation and let the \ac{cfb} control itself back to the original position.
During these maneuvers, we measure the estimated state of the \ac{cfb} and its motor commands.
The simulation then receives these motor commands along with the initial state and then generates the simulated trajectories.
We did not use these maneuvers to fit model parameters.
To also assess the effect of parameter uncertainties, we simulate the \ac{cfb} 4096 times with different inertia and amplify/reduce the thrust and torque of the motors. 
The randomly sampled parameters deviate at maximum  \SI{2}{\percent} from the identified ones.
Furthermore, we change the center of gravity by at maximum \SI{2}{\milli\meter}, which can be achieved by changing the matrix $L$ and apply velocity and angular noise per simulation step (\SI{4}{\milli\second}) of \SI{0.001}{\meter\per\second} and \SI{0.001}{\radian\per\second}.

\subsection{Results}
\cref{fig:evaluation_model:maneuvers} demonstrates that our model can capture the behavior of the \ac{cfb} in a horizon of \SI{200}{\milli\second}. 
Over time, the predictions of the model increasingly deviate from the real behavior.
We also noticed larger differences between measured and simulated velocities in z-direction especially during the second maneuver.

We suspect multiple sources for the remaining deviation of our model.
First, uncertainties have an increasingly large effect over the prediction horizon. 
Within this context, we observed that inaccuracies in the center of gravity position exert a particularly large influence.
Second, the measured values from the \ac{cfb} are taken from its Kalman filter (cf.~\cref{sec:background}) and might not necessarily represent the actual ground-truth values, they hence have their own errors and uncertainties. This is especially true for the velocity as there is no direct sensor for it (position, attitude and angular rates are measured by the \ac{mocap} and the \ac{imu}).
Third, our model neglects aerodynamics, particularly the complex aerodynamics of the motors, whose thrust changes during movement. Modeling these effects is non-trivial~\cite{bauersfeld2021neurobem,kaufmann2023champion} and represents an exciting topic for future work.
Fourth, during flight, we cannot measure the motor speeds and thus estimate their initial value, adding uncertainty.
Other potential factors include disturbances, like subtle air movements.

\begin{figure}[t]
    \vspace{3mm}
    \newcommand{\plotwidth}{0.45\linewidth}
    \newcommand{\plotheight}{0.38\linewidth}
    \centering
    \begin{tikzpicture}
        \begin{groupplot}[
            group style={
                group size=1 by 1,
                vertical sep=0.4cm,
                xlabels at=edge bottom,
                ylabels at=edge left,
            },
            width=\plotwidth,
            height=\plotheight,
            grid=major,
            xlabel style={font=\small},
            xtick={0.0, 0.1,0.2},
            ylabel style={font=\small},
            tick label style={font=\footnotesize},
            label style={font=\footnotesize},
            legend style={font=\footnotesize, at={(1.05,0.5)}, anchor=west, legend columns=2,nodes={scale=0.8, transform shape}},
            xmin=0, xmax=0.2,
            ymax=800, ymin=-800
        ]
        \nextgroupplot[
            xlabel={Time / \si{\second}},
            ylabel={Rate / \si{\degree\per\second}},
        ]
        \addplot[
            color=Dark2-A,
            mark=none,
            line width=1pt,
            dotted,
        ] table[
            x=time,
            y=gyro_x,
            col sep=comma
        ] {figure_data/cf21.csv};
        \addlegendentry{X (real)}
        \addplot[
            color=Dark2-B,
            mark=none,
            line width=1pt,
            dotted,
        ] table[
            x=time,
            y=gyro_y,
            col sep=comma
        ] {figure_data/cf21.csv};
        \addlegendentry{Y (real)}
        \addplot[
            color=Dark2-C,
            mark=none,
            line width=1pt,
            dotted,
        ] table[
            x=time,
            y=gyro_z,
            col sep=comma
        ] {figure_data/cf21.csv};
        \addlegendentry{Z (real)}
        \addplot[name path=gyroxmax, draw=none, forget plot] table[
            x=time,
            y=gyro_x_max,
            col sep=comma
        ] {figure_data/cf21_sim.csv};
        \addplot[name path=gyroxmin, draw=none, forget plot] table[
            x=time,
            y=gyro_x_min,
            col sep=comma
        ] {figure_data/cf21_sim.csv};
        \addplot[
            fill=Dark2-A, 
            fill opacity=0.3, 
            draw=none, forget plot
        ] fill between[
            of=gyroxmin and gyroxmax
        ];  
        \addplot[
            color=Dark2-A,
            mark=none,
            line width=1pt,
        ] table[
            x=time,
            y=gyro_x_mean,
            col sep=comma
        ] {figure_data/cf21_sim.csv};
        \addlegendentry{X (sim)}

        \addplot[name path=gyromymax, draw=none, forget plot] table[
            x=time,
            y=gyro_y_max,
            col sep=comma
        ] {figure_data/cf21_sim.csv};
        \addplot[name path=gyromymin, draw=none, forget plot] table[
            x=time,
            y=gyro_y_min,
            col sep=comma
        ] {figure_data/cf21_sim.csv};
        \addplot[
            fill=Dark2-B, 
            fill opacity=0.3, 
            draw=none, forget plot
        ] fill between[
            of=gyromymin and gyromymax
        ];  
        \addplot[
            color=Dark2-B,
            mark=none,
            line width=1pt,
        ] table[
            x=time,
            y=gyro_y_mean,
            col sep=comma
        ] {figure_data/cf21_sim.csv};
        \addlegendentry{Y (sim)}

        \addplot[name path=gyrozmax, draw=none, forget plot] table[
            x=time,
            y=gyro_z_max,
            col sep=comma
        ] {figure_data/cf21_sim.csv};
        \addplot[name path=gyrozmin, draw=none, forget plot] table[
            x=time,
            y=gyro_z_min,
            col sep=comma
        ] {figure_data/cf21_sim.csv};
        \addplot[
            fill=Dark2-C, 
            fill opacity=0.3, 
            draw=none, forget plot
        ] fill between[
            of=gyrozmin and gyrozmax
        ];  
        \addplot[
            color=Dark2-C,
            mark=none,
            line width=1pt,
        ] table[
            x=time,
            y=gyro_z_mean,
            col sep=comma
        ] {figure_data/cf21_sim.csv};
        \addlegendentry{Z (sim)}
       
        \end{groupplot}
    \end{tikzpicture}
    \caption{Comparison of hardware measurements (dotted lines) versus simulation (solid lines) of angular rates for the \ac{cf} 2.1. \capt{We use the model parameters used in PyBullet-drones~\cite{panerati2021learning} for baselines comparisons.}}
    \label{fig:evaluation_model:maneuverscf21}
\end{figure}
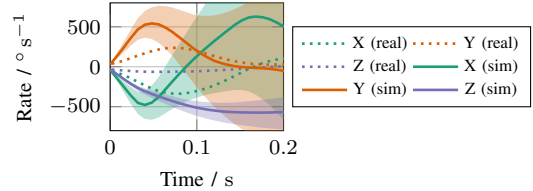

To contextualize the performance of our model, we compared the PyBullet-drones model for the \ac{cf} 2.1~\cite{panerati2021learning} against real flight data from a \ac{cf} 2.1 (\cref{fig:evaluation_model:maneuverscf21}). 
The results show that the PyBullet-drones model significantly deviates from the real behavior, demonstrating that our \ac{cfb} model achieves substantially higher accuracy than this established baseline.

While the presented model captures the fundamental behavior of the \ac{cfb}, some discrepancies remain. 
To address this, we investigate the effect of domain randomization on \ac{rl} performance in the following. 
The required degree of domain randomization for successful sim-to-real transfer serves as an additional metric for evaluating the accuracy of the dynamics model and the uncertainties in its key parameters.

\input{figures/compare_sim_real.tex}

\section{EVALUATION: SIMULATION-BASED REINFORCEMENT LEARNING}
\label{sec:evaluation_rl}

The previous section assessed the model's accuracy. In this section, we demonstrate the model's suitability for learning end-to-end \ac{nn} controllers by designing and evaluating a simulation-based \ac{rl} pipeline. Additionally, we use this pipeline to examine the impact of domain randomization, uncertainties applied to model parameters during training, which provides insights into the sim-to-real gap. 
A video accompanying this section is available at \href{http://tiny.cc/CFBVideo}{tiny.cc/CFBVideo}.

\subsection{Method}
\label{sec:evaluation_rl:method}

We train quadcopters on two distinct tasks. 
First, we focus on training a quadcopter to fly and maintain a specified target position. 
Second, we design a controller specifically for executing backflips.

\fakepar{Baselines} We compare the learned target controller with the firmware's PID and Mellinger controller~\cite{mellinger2011minimum,bitcraze2025crazyflie}, using parameters provided by the firmware.

\fakepar{Learning Algorithm}
We employ \ac{ppo}~\cite{schulman2017proximal} for learning, utilizing BRAX's \ac{ppo} implementation~\cite{brax2021github} and its interface to \ac{mjx} allowing the entire training to run on the GPU. 
To assess the sim-to-real gap, no dedicated fine-tuning is performed on hardware.

\fakepar{Controller}
The controller is a fully connected \ac{nn} with four hidden layers, each containing 32 neurons. 
Although the \ac{cfb} provides estimates of position, attitude, velocity and angular rates, it cannot measure motor rotation speeds $\Omega$.
Therefore, past estimates and actions are also included in the controller's input as a state representation.

\fakepar{Reward}
We use sums of triangular shaped rewards~\cite{cramer2025cheq}
\begin{equation}
\Lambda(e, e_\mathrm{max}) = \mathrm{relu}(1 - |e/e_\mathrm{max}|),
\end{equation}
where $e$ is a deviation from a target value (e.g. $p-p_\mathrm{target}$) and $e_\mathrm{max}$ is the maximum interval on which we want to give a reward.
This function helps to normalize the reward making the \ac{rl} training better conditioned.
We also add a large negative reward in case of a failure (e.g., crash to the ground or too high angular velocities).
The next paragraphs give the exact formulation for the specific controller.

\fakepar{Target Control} Goal of target control is to steer the quadcopter to the origin. 
We change the target position the quadcopter should fly to by subtracting it from the position estimate.
We use the following reward function to encourage proximity to the target position while penalizing high rotation speeds, angular velocities and generated torques (to ensure smooth maneuvers)
\begin{align}
    r(x) &= \Lambda(\twonorm{p}, \SI{10}{\meter}) + 0.05\Lambda(\twonorm{v}, \SI{10}{\meter\per\second})\nonumber \\
    &\opindent+ 0.15\Lambda(\twonorm{q-[1, 0, 0, 0]}, 2)\\
    &\opindent + 0.15\Lambda(\twonorm{\omega}, \SI{10}{\radian\per\second})\nonumber\\
    &\opindent + 0.02\Lambda(\twonorm{Lu}, \SI{0.01}{\newton\meter}) - 300\mathrm{fail}(x),
\end{align}
where $\mathrm{fail}$ is one when the quadcopter's velocity exceeds \SI{10}{\meter\per\second} or its angular velocity $\omega$ is above $\SI{30}{\radian\per\second}$.
In this case, we terminate the episode.

\fakepar{Backflips}
For backflips, we use a multi-phase approach similar to~\cite{antal2023backflipping}.
First, we use the controller from the last section to accelerate the quadcopter upwards and then switch to a second \ac{nn} controller (called \flipcontroller{} in the following) that controls the quadcopter to have a constant angular rate around its x-axis and a given velocity along its local frame's z-axis.
After some time, we switch back to the stabilizing controller that catches the quadcopter and steers its back towards its starting position.
The possibility to vary angular rate and velocity setpoints $v_\mathrm{target},\dot{\theta}_\mathrm{target}$ and the time points on which we switch the controllers allows us to change the behavior of the drone during the backflip, generating looping of different sizes and speeds.

The \ac{nn} controller for the second phase is trained via the following reward
\begin{align}
    r_\mathrm{flip}(x) &= 0.6\Lambda(\twonorm{R(q)^Tv - \begin{bmatrix}0 & 0 & v_\mathrm{target}\end{bmatrix}^T}, \SI{20}{\meter\per\second})\nonumber \\
    &\opindent+ 0.15\Lambda(\twonorm{R(q)[:, 0] - \begin{bmatrix}0&1&0\end{bmatrix}^T}, 3.0)\nonumber\\
    &\opindent + 1.0\Lambda(\twonorm{\omega - \begin{bmatrix}\dot{\theta}_\mathrm{target}&0&0\end{bmatrix}}, \SI{0.01}{\newton\meter})\nonumber\\
    &\opindent- 300\mathrm{fail}(x),
\end{align}
where the first part rewards keeping the velocity, the second keeping the quadcopter's attitude close to its rotation axis and the third rewarding controlling it to the pitch rate.

\fakepar{Domain Randomization}
We randomized key model parameters: mass ($\pm\SI{10}{\percent}$), inertia and motor dynamics ($\pm\SI{20}{\percent}$), motor thrusts and torques scalings ($\pm\SI{20}{\percent}$) and center of gravity position ($\pm\SI{1}{\centi\meter}$), sampling from uniform distributions across these intervals.
Additionally, we introduced random offsets in attitude measurements ($\pm\SI{15}{\degree}$) to simulate real-world misalignment between the drone's local coordinate frame, the global \ac{mocap} frame and the direction of gravity.

We quantify the effect of domain randomization in \cref{sec:evaluation_rl:domain_randomization} by training \ac{nn} controllers with varying magnitudes of randomization, defined as a factor for the interval size.
A magnitude of 1.0 corresponds to the full intervals specified above, while 0.5 and 0.0 correspond to half-intervals and no randomization, respectively.

\subsection{Evaluation: Target Control}
\begin{figure}[t]
    \vspace{3mm}
    \newcommand{\plotwidth}{0.99\linewidth}
    \newcommand{\plotheight}{0.3\linewidth}
    \centering
    \tikzset{external/export next=false}
    \begin{tikzpicture}
        \begin{groupplot}[
            group style={
                group size=1 by 2,
                xlabels at=edge bottom,
                ylabels at=edge left,
                vertical sep=0.05\textwidth,
            },
            width=\plotwidth,
            height=\plotheight,
            xlabel style={font=\footnotesize},
            ylabel style={font=\footnotesize},
            tick label style={font=\scriptsize},
            legend style={font=\footnotesize, at={(0.5,1.05)}, anchor=south, legend columns=4,nodes={scale=0.8, transform shape}},
            grid=major,
            grid style={gray!30},
            xmin=-0.01, xmax=35.0,
        ]

        \nextgroupplot[
            ylabel={Position / \si{\meter}},
            ymin=-3.2, ymax=3.2
        ]
        \addplot[color=Dark2-A, line width=1pt,] table[x=time, y=x, col sep=comma] {figure_data/target_control2.csv};
        \addlegendentry{X}
        \addplot[color=Dark2-B, line width=1pt,] table[x=time, y=y, col sep=comma] {figure_data/target_control2.csv};
        \addlegendentry{Y}
        \addplot[color=Dark2-C, line width=1pt,] table[x=time, y=z, col sep=comma] {figure_data/target_control2.csv};
        \addlegendentry{Z}
        \addplot[color=black, line width=1.5pt, dashed] coordinates {
            (0.0, -5.0)
            (1.0, -5.0)
        };
        \addlegendentry{Setpoints}
        \addplot[color=Dark2-A, line width=1.5pt, dashed, forget plot] coordinates {
            (0.0, 0.0)
            (8.0, 0.0)
            (8.0, 3.0)
            (16.0, 3.0)
            (16.0, -3.0)
            (24.0, -3.0)
            (24.0, 3.0)
            (32.0, 3.0)
            (32.0, 0.0)
        };
        \addplot[color=Dark2-B, line width=1.5pt, dashed, forget plot] coordinates {
            (0.0, 0.0)
            (35.0, 0.0)
        }; 
        
        \addplot[color=Dark2-C, line width=1.5pt, dashed, forget plot] coordinates {
            (0.0, 1.0)
            (35.0, 1.0)
        };   

        \addplot[color=Dark2-C, line width=1.5pt, dashed, forget plot] coordinates {
            (0.0, 1.0)
            (35.0, 1.0)
        };

        \nextgroupplot[
            ylabel={Velocity / \si{\meter\per\second}},
            ymin=-5.9, ymax=5.9,
            xlabel={Time / \si{\second}}
        ]
        \addplot[color=Dark2-A, line width=1pt,] table[x=time, y=vx, col sep=comma] {figure_data/target_control2.csv};
        \addplot[color=Dark2-B, line width=1pt,] table[x=time, y=vy, col sep=comma] {figure_data/target_control2.csv};
        \addplot[color=Dark2-C, line width=1pt,] table[x=time, y=vz, col sep=comma] {figure_data/target_control2.csv};

        \end{groupplot}
    \end{tikzpicture}
    \caption{Target controller flying back and forth between two points \SI{6}{\meter} apart. \capt{The \ac{nn} controller guides the \ac{cfb} to its target with an accuracy of a few \si{\centi\meter} (cf. \cref{tab:compare_controllers}), achieving velocities of approximately \SI{5}{\meter\per\second} during the maneuver.}}
    \label{fig:target_control_rl}
\end{figure}
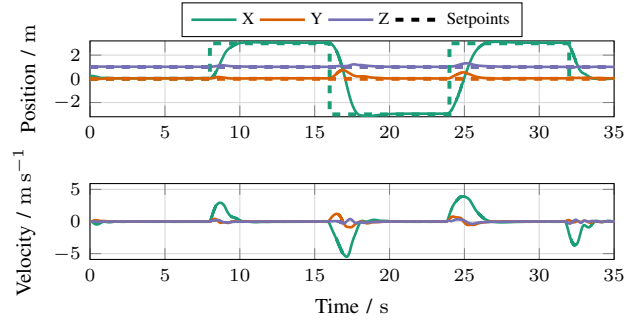
\cref{fig:target_control_rl} illustrates a maneuver executed by the \ac{cfb}, flying between two positions \SI{6}{\meter} away. The quadcopter successfully reaches its target within \SI{2}{\second}, achieving a maximum velocity of around \SI{5}{\meter\per\second}. 

In \cref{tab:compare_controllers}, we compare the learned controller for different domain randomization magnitudes with the PID and Mellinger controllers from the firmware. We conducted a hover test for \SI{6}{\second}, assessing mean distance to target position and motor command standard deviation to evaluate motor command smoothness. 

The PID has the best positional accuracy, followed by the \ac{nn} controller with domain randomization magnitude 1.0 and then the Mellinger controller.
The \ac{nn} controller performs slightly better in motor command smoothness. 

These results demonstrate the \ac{rl} pipeline's ability to learn a controller using our model, whose stationary behavior is comparable to the firmware's controllers.
We want to note that we used the PID and Mellinger controller's standard parameters and tuning of these controllers could potentially improve their performance.

\input{figures/domain_randomization.tex}

\begin{table}[t]
    \vspace{1mm}
\centering
\fontsize{9}{11}\selectfont
\caption{Comparison of the learned \ac{nn} controllers for different domain randomization magnitudes (denoted in parentheses, cf. \cref{sec:evaluation_rl:method}) against the PID and Mellinger controllers from the firmware.
\capt{The first column is the mean differences from setpoint position measured for \SI{5}{s} and the second mean of motor command standard deviations (std) over the four motors.}}
\label{tab:compare_controllers}
\begin{tabular}{l S[table-format=2.2] S[table-format=1.4]}
    \hline
    Controller      & {Mean Distance (mm)} & {Motor command std} \\
    \hline
    \ac{nn} (0) & 284.76 & 0.0203 \\
    \ac{nn} (0.5) & 100.07 & 0.0245 \\
    \ac{nn} (1.0) & 43.16 &0,0197 \\
    \ac{nn} (1.5) & 71.39 & 0.0548 \\
    PID       & 27.81                & 0.0226             \\
    Mellinger & 54.68                & 0.0394             \\
    \hline
\end{tabular}
\end{table}

\subsection{Evaluation: Backflips}
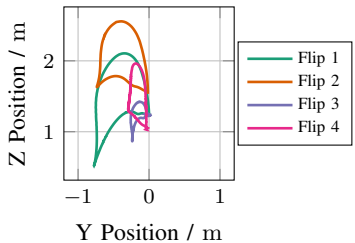
\begin{figure}[t]
    \vspace{3mm}
    \tikzset{external/export next=false}
    \newcommand{\plotwidth}{0.45\linewidth}
    \newcommand{\plotheight}{0.45\linewidth}
    \centering
    
    \begin{tikzpicture}
        \begin{axis}[
            width=\plotwidth,
            height=\plotheight,
            grid=major,
            xlabel={Y Position / \si{\meter}},
            ylabel={Z Position / \si{\meter}},
            xlabel style={font=\small},
            ylabel style={font=\small},
            tick label style={font=\footnotesize},
            axis equal image,
            legend style={
                at={(1.02,0.5)},
                anchor=west,
                font=\footnotesize, legend columns=1,nodes={scale=0.8, transform shape}},
            xmin=-1.2, xmax=1.2,
        ]

        \addplot[
            color=Dark2-A,
            mark=none,
            line width=1.0pt,
        ] table[
            x=y,
            y=z,
            col sep=comma
        ] {figure_data/bf2.csv};
        \addlegendentry{Flip 1}

        \addplot[
            color=Dark2-B,
            mark=none,
            line width=1.0pt,
        ] table[
            x=y,
            y=z,
            col sep=comma
        ] {figure_data/bf3.csv};
        \addlegendentry{Flip 2}

        \addplot[
            color=Dark2-C,
            mark=none,
            line width=1.0pt,
        ] table[
            x=y,
            y=z,
            col sep=comma
        ] {figure_data/bf5.csv};
        \addlegendentry{Flip 3}
        
        Flip 3
        \addplot[
            color=Dark2-D,
            mark=none,
            line width=1.0pt,
        ] table[
            x=y,
            y=z,
            col sep=comma
        ] {figure_data/bf4.csv};
        \addlegendentry{Flip 4}
        
        \end{axis}
    \end{tikzpicture}
    
    \caption{Trajectory in YZ plane for four flip maneuvers. \capt{Depending on the hyperparameters of the maneuver (more specifically duration of the liftoff and roll rate), the shape changes.}}
    \label{fig:backflip}
\end{figure}

\fakepar{Single Backflips}
\cref{fig:backflip} illustrates various backflips executed by the \ac{nn} controllers.
The shape of each backflip varies according to the setpoint velocity and roll rate.
As the \ac{cfb} cannot control its velocity when inverted, the velocity is not kept exactly.
However, the adjustable setpoints still allow the users to customize the backflips to their preferences.

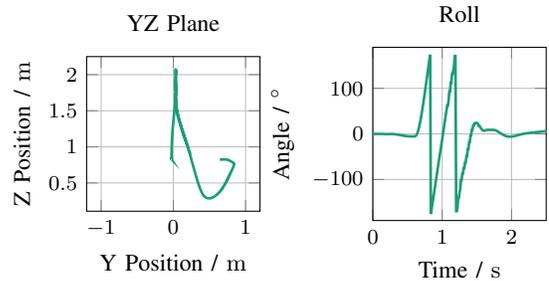
\begin{figure}[t]
    \vspace{2mm}
    \tikzset{external/export next=false}
    \newcommand{\plotwidth}{0.45\linewidth}
    \newcommand{\plotheight}{0.45\linewidth}
    \newcommand{\attitudeplotwidth}{0.35\linewidth}
    \newcommand{\attitudeplotheight}{0.32\linewidth}
    \centering
        \centering
        \begin{tikzpicture}
            \begin{groupplot}[
                group style={
                    group size=2 by 1,
                    horizontal sep=1.5cm,
                    xlabels at=edge bottom,
                    ylabels at=edge left,
                },
                width=\plotwidth,
                height=\plotheight,
                grid=major,
                xlabel style={font=\small},
                ylabel style={font=\small},
                tick label style={font=\footnotesize},
            ]
            
            \nextgroupplot[
                xlabel={Y Position / \si{\meter}},
                ylabel={Z Position / \si{\meter}},
                title={YZ Plane},
                title style={font=\small},
                xmin=-1.2, xmax=1.2,
                axis equal image,
            ]
            \addplot[
                color=Dark2-A,
                mark=none,
                line width=1.0pt,
            ] table[
                x=y,
                y=z,
                col sep=comma
            ] {figure_data/double_flip.csv};

            \nextgroupplot[
                xlabel={Time / \si{\second}},
                ylabel={Angle / \si{\degree}},
                title={Roll},
                title style={font=\small},
                ymin=-190, ymax=190,
                xmin=0, xmax=2.5,
            ]
            \addplot[
                color=Dark2-A,
                mark=none,
                line width=1pt,
            ] table[
                x=time,
                y=roll,
                col sep=comma
            ] {figure_data/double_flip.csv};    
            \end{groupplot}
        \end{tikzpicture}
    \caption{Double backflip maneuver showing trajectory in YZ plane and roll angle over time.}
    \label{fig:double_backflip}
\end{figure}

\fakepar{Double Backflips}
\cref{fig:double_backflip} depicts the trajectory of a \ac{cfb} executing a backflip with two rotations. 
This maneuver is performed requiring only $\SI{1.8}{\meter}$ movement in z-direction. 
The quadcopter reduces its rotational velocity from \SI{1000}{\deg\per\second} in \SI{0.35}{\second} and its vertical velocity from \SI{3.5}{\meter\per\second} to zero in \SI{0.6}{\second}, demonstrating the \ac{cfb}'s capacity for agile, high-thrust maneuvers.

In summary, \ac{rl} using the presented model enables the \ac{cfb} to execute high-performance aerial acrobatics.

\subsection{Effect of Domain Randomization}
\label{sec:evaluation_rl:domain_randomization}

To assess the impact of domain randomization, we trained the target controller using various randomization magnitudes. 
The results of these experiments are presented in \cref{tab:compare_controllers} and \cref{fig:domain_randomization}.

Low magnitudes of domain randomization reduce positional accuracy, primarily in the z-direction. 
Nevertheless, even with low or no randomization, the drone maintains a level and stable flight. 
Conversely, excessively high domain randomization significantly degrades controller performance, as the increased demand for robustness comes at the expense of optimal performance.
These findings suggest that the rotational dynamics are less sensitive to low domain randomization than the vertical dynamics.
We hypothesize that this is due to slight discrepancies in mass and thrust between the model and the real quadcopter, combined with reduced controller robustness to these parameters when trained with limited randomization.

We note that the reward functions were not explicitly tuned for different randomization magnitudes. Such tuning could potentially improve performance. Alternatively, the loss of accuracy in the z-direction might be mitigated by removing mass compensation from the \ac{nn} controller and instead compensating for it with a constant offset to the motor commands.

In summary, our experiments demonstrate that the presented model is accurate enough to successfully train \ac{nn} controllers in simulation that can be successfully deployed on a real \ac{cfb}. To enhance robustness against model-to-real-world discrepancies, we recommend domain randomization of $\pm\SI{10}{\percent}$ to $\pm\SI{20}{\percent}$ for mass, inertia, motor model and thrust/torque scaling. 
For controllers designed primarily for attitude control, where translational dynamics are less critical, a reduced amount or even the omission of domain randomization may be sufficient.

\section{CONCLUSION AND OUTLOOK}

This work derived a dynamics model of the recently released \ac{cfb}, to support future research with this new version of the popular Crazyflie platform. Specifically, we identified the parameters of the dynamics model and provided procedures to quickly re-identify the inertias when payloads are added or removed.
Evaluations of the model's prediction capabilities demonstrated that the model is accurate with minor discrepancies due to unmodeled effects such as aerodynamics and observer dynamics. 
Additionally, we demonstrated the model's effectiveness by training end-to-end \ac{nn} controllers through \ac{rl} that transfer from simulation to real world. 
Our experiments revealed that domain randomization between \SI{10}{\percent}--\SI{20}{\percent} effectively bridges the sim-to-real gap for \ac{rl}, allowing end-to-end \ac{nn} controllers that generalize from simulation to real-world deployment on the \ac{cfb}.

To our knowledge, this is the first openly available model of the new \ac{cfb}, effective for common tasks like \ac{rl}. 
Future work could extend this model by identifying the \ac{cfb}'s aerodynamics, for instance using methods like those in~\cite{hanover2024autonomous}, which is particularly relevant for high-speed applications like quadcopter racing.
Another interesting direction is modeling the aerodynamic interactions between multiple \ac{cfb}, as explored in~\cite{shi2021neural}, to facilitate the development of high-performance swarms.

\addtolength{\textheight}{-0cm}   





\section*{ACKNOWLEDGMENT}
We thank Emma Cramer, Bernd Frauenknecht, Henrik Hose, David Stenger, Devdutt Subhasish, Khaled Wahba and Lukas Wildberger for helpful discussions and valuable feedback.

\bibliographystyle{IEEEtran}
\bibliography{references}

\end{document}